\pdfoutput=1
\documentclass[runningheads]{llncs}
\usepackage{url}
\usepackage{booktabs}
\usepackage{graphicx}
\usepackage{hyperref}

\usepackage{bbding}
\usepackage{algorithm}
\usepackage[fleqn]{amsmath}
\usepackage{mathtools}
\usepackage{amsfonts,amssymb}
\usepackage{pifont}
\usepackage[nocompress]{cite}
\usepackage{booktabs} 
\usepackage[normalem]{ulem}
\usepackage{graphicx}
\usepackage{amsmath}
\usepackage{amssymb}
\usepackage{color}
\usepackage{algorithmic,algorithm}
\usepackage{hyperref}
\usepackage{multirow}
\usepackage{mathtools}
\usepackage{enumitem}
\useunder{\uline}{\ul}{}
\usepackage{threeparttable}
\usepackage{todonotes}
\usepackage{adjustbox}
\usepackage{caption}
\usepackage{subcaption}

\usepackage{wrapfig,lipsum} 
\definecolor{applegreen}{rgb}{0.55, 0.71, 0.0}

\newcommand{\FTN}{FTN}\newcommand{\STN}{STN}

\newcommand{\Reals}{\mathbb{R}}

\newcommand{\dataset}{\mathit{D}}

\newcommand{\Ei}{\mathit{E}_\theta}
\newcommand{\Di}{\mathcal{D}_{\phi_{i}}}
\newcommand{\Ds}{\mathcal{D}_{\phi_{s}}}
\newcommand{\C}{\mathcal{C}_{\psi}}

\newcommand{\filter}{\mathcal{H}}
\newcommand{\gdp}{\G_\mathit{dp}}
\newcommand{\gch}{\G_\mathit{ch}}
\newcommand{\gsp}{\G_\mathit{sp}}

\newcommand{\EX}{\mathbb{E}}\newcommand{\G}{\mathcal{G}}

\newcommand{\image}{\mathbf{x}}
\newcommand{\segmentation}{\mathbf{y}}

\newcommand{\prediction}{\mathbf{p}}
\newcommand{\mask}{\mathbf{m}}

\newcommand{\z}{\mathbf{z}}
\newcommand{\zi}{\mathbf{z}_i}
\newcommand{\zs}{\mathbf{z}_s}
\newcommand{\g}{\mathbf{g}}

\newcommand{\zero}{\mathbf{0}}
\newcommand{\one}{\mathbf{1}}

\newcommand{\Loss}{\mathcal{L}}
\newcommand{\SegLoss}{\mathcal{L}_\mathit{seg}}
\newcommand{\ShapeLoss}{\mathcal{L}_{\mathit{shp}}}
\newcommand{\RecLoss}{\mathcal{L}_{\mathit{rec}}}

\title{Cooperative Training and Latent Space Data Augmentation for Robust Medical Image Segmentation}
\author{Chen Chen\inst{1}(\Envelope),  Kerstin Hammernik\inst{1,2}, Cheng Ouyang\inst{1}, Chen Qin\inst{3},  Wenjia Bai\inst{4,5}, Daniel Rueckert\inst{1,2}}
\authorrunning{C. Chen et al.}
\institute{BioMedIA Group, Department of Computing, Imperial College London, UK
\and Klinikum rechts der Isar, Technical University of Munich, Germany
\and Institute for Digital Communications, University of Edinburgh, UK
\and Data Science Institute, Imperial College London, UK
\and Department of Brain Sciences, Imperial College London, UK
\email{chen.chen15@imperial.ac.uk}}

\titlerunning{Cooperative Training and Latent Space Data Augmentation}

\begin{document}
\maketitle

\begin{abstract}
Deep learning-based segmentation methods are vulnerable to unforeseen data distribution shifts during deployment, e.g. change of image appearances or contrasts caused by different scanners, unexpected imaging artifacts etc. In this paper, we present a cooperative framework for training image segmentation models and a latent space augmentation method for generating hard examples. Both contributions improve model generalization and robustness with limited data. The cooperative training framework consists of a fast-thinking network (\FTN) and a slow-thinking network (\STN). The \FTN\ learns decoupled image features and shape features for image reconstruction and segmentation tasks. The \STN\ learns shape priors for segmentation correction and refinement. The two networks are trained in a cooperative manner. The latent space augmentation generates challenging examples for training by masking the decoupled latent space in both channel-wise and spatial-wise manners. We performed extensive experiments on public cardiac imaging datasets. Using only 10 subjects from a \emph{single} site for training, we demonstrated improved cross-site segmentation performance, and increased robustness against various unforeseen imaging artifacts compared to strong baseline methods. Particularly, cooperative training with latent space data augmentation yields 15\% improvement in terms of average Dice score when compared to a standard training method.

\end{abstract}

 \section{Introduction}
\label{SEC:introduction}
Segmenting anatomical structures from medical images is an important step for diagnosis, treatment planning and clinical research. In recent years, deep convolutional neural networks (CNNs) have been widely adopted to automate the segmentation procedure~\cite{Shen_2017_Review,litjens_2017_survey}. However, a major obstacle for deploying deep learning-based methods to real-world applications is  domain shift during clinical deployment, which includes changes of image appearance and contrasts across medical centers and scanners as well as various imaging artefacts. Recent works on domain generalization provide a promising direction to address this issue~\cite{Dou_2019_NIPS,albuquerque_2020_Arxiv_improving,chattopadhyay_2020_ECCV_learning_to_balance,Wang_2020_ECCV,Shankar_2018_ICLR}. A majority of them require training data from \emph{multiple} domains to learn domain-invariant features for segmentation. Multi-domain datasets, however, may not always be feasible due to data privacy concerns and collection costs. Learning robust networks from single-domain data and limited data is of great practical value for medical imaging research.

In this work, we propose a novel cooperative training framework for learning a robust segmentation network
from \emph{single-domain} data. We make the following contributions. (1) First, to improve model performance on unseen domains, we design a cooperative training framework where two networks collaborate in both training and testing. This is inspired by the two-system model in human behaviour sciences~\cite{daniel_2017_thinking}, where a fast-thinking system makes intuitive judgment and a slow-thinking system corrects it with logical inference. Such a collaboration is essential for humans to deal with unfamiliar situations. In our framework, a fast-thinking network (\FTN) aims to understand the context of images and extracts task-related image and shape features for an initial segmentation. Subsequently, a slow-thinking network (\STN) refines the initial segmentation according to a learned shape prior. (2) We introduce a latent space data augmentation (DA) method, which performs channel-wise and spacial-wise masking for the latent code learnt from \FTN\ in random and targeted fashions. Reconstructing images with masked latent codes generates a diverse set of challenging images and corrupted segmentation maps to reinforce the training of both networks. Experimental results on cardiac imaging datasets show the cooperative training mechanism with generated challenging examples can effectively enhance \FTN's segmentation capacity and \STN's shape correction ability, leading to more robust segmentation. (3) The proposed method alleviates the need for multi-domain data and expertise for data augmentation, making it applicable to a wide range of applications.

\noindent\textbf{Related Work.}
Our work is conceptually related to DA, multi-task learning (MTL).
\emph{a) DA} applies transformations or perturbations to improve the diversity of training data, which is effective for improving model generalization~\cite{Shorten_2019_Bigdata}. A large number of the works focuses on image-space DA, including both intensity and geometric transformation functions~\cite{Zhang_2020_TMI,chen_2020_frontiers} and patch-wise perturbations~\cite{Devries_2017_Cutout,Raphael_2019_Arxiv_PatchGaussian,Zhou_2019_MICCAI_ModelsGenesis,Zhou_2021_Media_ModelsGenesis}. Adversarial DA has also been explored, which takes the segmentation network into account and generates adversarial examples that can fool the network~\cite{Miyato_2018_PAMI_VAT,chen_2020_MICCAI,Zhang_2020_DADA,Zhao_2020_NIPS}. A major novelty of our work is that we perform DA in the latent space. The latent space contains abstract representation of both image and shape features and challenging examples can be generated by manipulating this space. Different from existing latent DA methods used in metric learning~\cite{Zheng_2019_CVPR}, our method is based on feature masking rather than feature interpolation and thus does not require paired images from the same/different categories to generate synthetic data. To the best of our knowledge, our work is the first to explore latent space DA for robust segmentation with single domain data.
\emph{b) MTL} is extremely beneficial when training data is limited~\cite{zhang_2017_survey_MTL,Zhou_2021_Media_ModelsGenesis,Zhou_2019_MICCAI_ModelsGenesis}. MTL enhances network capacity by encouraging the learning of common semantic features across various tasks.  Several methods consist of two stages of segmentation: a first network for coarse segmentation from images and a second network for refinement, where two networks are trained independently~\cite{Larrazabal_2019_MICCAI_DAE,painchaud_2019_MICCAI_cardiac}. For example, in~\cite{Larrazabal_2019_MICCAI_DAE}, manually designed functions are used to generate poor segmentation and a denoising autoencoder is independently trained for segmentation refinement. Another novelty of our work is that we seek the mutual benefits of a segmentation network and a denoising network by training them cooperatively, using hard examples constructed from latent space.

 \section{Methodology}
\label{SEC:methodology}
Given a training dataset from \emph{one, single} domain $\dataset_{tr} = \{(\image_i,\segmentation_{i})\}_{i=1}^{n}$, with pairs of images $\image_i \in \Reals^{H \times W}$ and one-hot encoded $C$-class label maps $\segmentation_i \in \{0,1\}^{H\times W \times C}$ as ground truth (GT), our goal is to learn a robust segmentation network across various `unseen' domains with different image appearance and/or quality. Here, $H,W$ denote image height and width, respectively.

\subsection{Overview of the framework}
\setlength\intextsep{0pt}
\begin{figure}[t]
    \centering
    \includegraphics[width=\textwidth]{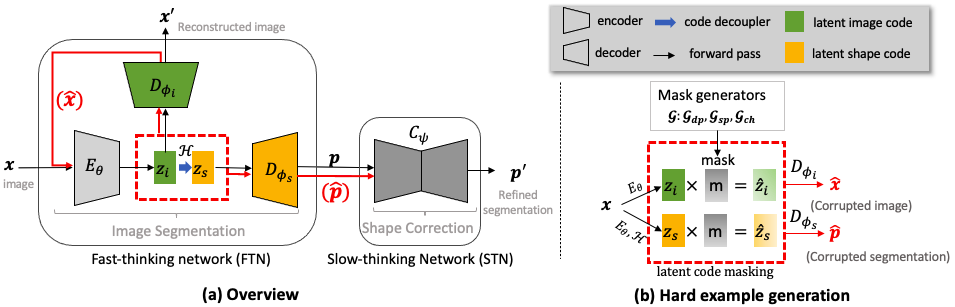}
    \caption{(a) The proposed cooperative training framework, which consists of a fast-thinking network (\FTN) and a slow-thinking network (\STN).
        (b) Hard example generation in latent space. Latent code masking is performed for generating both corrupted images and segmentations for cooperative training.}
            \label{fig:introduction}
\end{figure}
\setlength\intextsep{0pt}
 
An overview of the proposed framework is illustrated in Fig.~\ref{fig:introduction} (a). At a high level, our framework consists of a fast-thinking network (\FTN) and a slow-thinking network (\STN). Given an image $\image$, the \FTN\ extracts task-specific shape features $\zs$ to perform the segmentation task and image contextual features $\zi$ to perform the image reconstruction task. This network consists of a shared encoder $\Ei$, a feature decoupler $\filter$ and two task-specific decoders $\Ds$ and $\Di$ for image segmentation and reconstruction tasks. We apply the latent code decoupler $\filter$ to $\zi$, so that task-unrelated information (e.g. image texture information, brightness) is deactivated in $\zs$. This encourages a sparse latent code $\zs$, which is beneficial for model robustness~\cite{Tishby_2000_information}. $\filter$ employs a stack of two convolutional layers followed by a ReLU activation function. Please find Supple. Fig.5 for more details.
\STN\ is a denoising autoencoder network $\C$, which corrects the segmentation predicted by \FTN\ by using a learned shape prior encoded in $\C$. At inference time, we first employ \FTN\ to perform fast segmentation for a given image $\image$: $\prediction=\Ds(\filter(\Ei(\image)))$, and then \STN\ to refine the prediction for improved segmentation quality: $\prediction'=\C(\prediction)$.
 
\subsection{Standard training}
To train the two networks, we propose a standard approach which jointly trains the three encoder-decoder pairs with a supervised multi-task loss function for image reconstruction $\RecLoss$, image segmentation $\SegLoss$ and shape correction $\ShapeLoss$. The loss is defined as:\begin{small}
\begin{equation}
    \Loss_{\mathit{std}} =\EX_{(\image,\segmentation)\in \dataset_{tr}}[ \RecLoss(\image',\image) + \\
    \SegLoss(\prediction,\segmentation)+\\
    \ShapeLoss(\prediction',\segmentation)+\ShapeLoss(\segmentation',\segmentation)],
\end{equation}
\end{small}
where $\RecLoss$ is the mean squared error (MSE) between the original input image $\image$ and the reconstructed image $\image'=\Di(\Ei(\image))$, $\SegLoss$ and $\ShapeLoss$ are cross-entropy loss functions between ground truth $\segmentation$ and predicted segmentation. The predicted segmentation can be the initial prediction $\prediction=\Ds(\filter(\Ei(\image)))$, or the reconstructed prediction $\prediction' = \C(\prediction)$ or reconstructed ground-truth map $\segmentation'= \C(\segmentation)$. Different from $\SegLoss$, optimizing $\ShapeLoss(\prediction',\segmentation)$ will trigger gradient flows from \STN\ to \FTN. This allows \STN\ to transfer shape knowledge to \FTN\ to improve model generalizability.

\subsection{Latent space data augmentation for hard example generation}
\label{SEC:latent_code_masking}
Standard training is likely to suffer from over-fitting when training data is limited. To solve this problem, a novel latent space DA method is proposed which allows \FTN\ to automatically construct hard examples. As shown in Fig.~\ref{fig:introduction}(b), the proposed method requires a mask generator $\G$ to produce a mask $\mask$ on the latent code $\z$. The masked latent code $\hat\z = \z \cdot \mask$ is then fed to the decoders to reconstruct a corrupted image $\hat\image=\Di(\hat\z_i)$ and segmentation $\hat\prediction=\Ds(\hat\z_s)$. Here, $\cdot$ denotes element-wise multiplication. In our work, we use latent code masking for data augmentation. This differs from existing latent code dropout techniques for explicit regularization~\cite{Huang_2020_ECCV_Self_challenging,Tompson_2015_CVPR_SpatialDropout}. By dynamically masking the latent code, the proposed method can generate samples with a wide diversity of image appearances and segmentations,  which are not bound to specific image transformation or corruption functions. Below we introduce three latent-code masking schemes: random dropout $\gdp$, and two targeted masking schemes, channel-wise targeted mask generation $\gch$ and spatial-wise targeted mask generation $\gsp$. \\

\noindent\textbf{(1) Random Masking with Dropout}
A na\"ive approach for latent code masking is random channel-wise dropout~\cite{Tompson_2015_CVPR_SpatialDropout}, which is an enhanced version of the original dropout method. An entire channel of the latent code can be masked with all zeros at a probability of $p$ at training. Mathematically, this can be viewed as sampling a mask from a Bernoulli distribution:
\begin{small}
\begin{equation}
\label{Eq:dropout}
\gdp(\mask^{(i)};p) =
 \left\{
\begin{array}{ll}
       p  & \mask^{(i)} =\zero \in \Reals^{h \times w} \\
       1-p & \mask^{(i)} =\one  \in \Reals^{h \times w}; \;
\end{array} 
\right. \forall i \in {1,...,c}.
\end{equation}
\end{small}
The masked code at $i$-th channel is obtained via $\hat{\z}^{(i)} = \z^{(i)} \cdot \mask^{(i)}$. In the following, we will use i-j-k to denote the three coordinates of latent code $\z \in \Reals^{c \times h \times w}$. \\

\noindent\textbf{(2) Targeted Masking}
Inspired by the recent success on latent code masking for domain generalized image classification algorithm~\cite{Huang_2020_ECCV_Self_challenging}, we propose targeted latent code masking schemes which takes gradients as a clue to identify `salient' features to mask. Following the common practice in adversarial DA~\cite{Madry_2017_PGDattack,Goodfellow_2015_FGSM}, we take task-specific losses (image reconstruction loss and image segmentation loss) to calculate the gradients $\g_{\zi}$, $\g_{\zs}$ for $\zi$ and $\zs$ respectively, formulated as: 
$
\g_{\zi} = \nabla_{\zi}\RecLoss{(\Di(\zi),\image})
$,
$
\g_{\zs} = \nabla_{\zs}\SegLoss(\Ds(\zs),\segmentation)
$.  By ranking the values of task-specific gradients, we can identify most predictive elements in the latent space to attack. We hypothesize that the elements with high response to task-specific loss functions are leading causes to performance drop under unforeseen domain shifts. We therefore focus on attacking these primary elements to simulate strong data distribution shifts. Two types of targeted masking are implemented, which mask features in latent code $\z$ along the channel dimension and spatial dimension.\\
\textbf{a) channel-wise mask generator:
} 
\begin{small}
\begin{equation}
\label{Eq:channel}
\gch(\mask^{(i)};\g_\z, p) =
 \left\{
\begin{array}{ll}
      \mask^{(i)} = a \one\in \Reals^{h \times w} & \text{if}\; \EX[\g_\z^{(i)}] \geq z^{ch}_p \\
       \mask^{(i)} = \one\in \Reals^{h \times w} & \text{if}\; \EX[\g_\z^{(i)}] < z^{ch}_p;
\end{array} 
\right. \forall i \in {1,...,c}, 
\end{equation}
\end{small}
\textbf{b) spatial-wise mask generator:}
\begin{small}
\begin{equation}
\label{Eq:spatial}
\gsp(\mask^{(j,k)};\g_\z,p) =
 \left\{
\begin{array}{ll}
      \mask^{(j,k)} = a \one \in \Reals^{c}  & \text{if}\; \EX[\g_\z^{(j,k)}] \geq z_p^{sp} \\
       \mask^{(j,k)} = \one \in \Reals^{c} & \text{if}\; \EX[\g_\z^{(j,k)}] < z^{sp}_p;
\end{array} 
\right. \forall j \in [1,h], \forall k \in [1,w]. 
\end{equation}
\end{small}
Thresholds $z_p^{ch}, z_p^{sp} \in \Reals$ are top $p$-th value across the channel means and spatial means.
$a$ is an annealing factor randomly sampled from (0,0.5) to create soft masks. Compared to hard-masking ($a$=0), soft-masking generates more diverse corrupted data (Supple. Fig.1). Channel-wise masked code at $i$-th channel is obtained via $\hat{\z}^{(i)} = \z^{(i)} \cdot \mask^{(i)}$.
Spatial-wise masked code at $(j,k)$ position is obtained via $\hat{\z}^{(j,k)} = \z^{(j,k)} \cdot \mask^{(j,k)}$.

\subsection{Cooperative training}
During training, we randomly apply one of the three mask generators to both  $\zi$, $\zs$. This process generates a rich set of corrupted images $\hat\image$ and segmentations $\hat\prediction$ on-the-fly. It allows us to train our dual-network on three hard example pairs, i.e. corrupted images-clean images 
$({\hat{\image},\image})$, corrupted images-GT
$({\hat{\image},\segmentation})$,
corrupted prediction-GT 
$({\hat{\prediction},\segmentation})$. The final loss for the proposed cooperative training method is a combination of losses defined on easy examples and hard examples:
$\Loss_{\mathit{cooperative}} = \Loss_{\mathit{std}} +\Loss_{\mathit{hard}}$, where $\Loss_{\mathit{hard}}$ is defined as:
\begin{small}
\begin{equation}
    \Loss_{\mathit{hard}} =\EX_{\hat\image,\hat\prediction,\image,\segmentation}[ \RecLoss( \Di(\Ei(\hat{\image})),\image)+\SegLoss{(\bar{\prediction},\segmentation)}+\ShapeLoss{(\C(\hat\prediction),\segmentation)}+\ShapeLoss{(\C(\bar{\prediction}),\segmentation)}].
\end{equation}
\end{small}
Here, $\bar\prediction = \Di(\filter(\Ei(\hat\image)))$ is \FTN's predicted segmentation on $\hat\image$. 

\section{Experiments and Results}
\label{SEC: experiment}
\noindent\textbf{Datasets.} To evaluate the efficacy of the proposed method, we apply it to the cardiac image segmentation task to segment the left ventricle cavity, left ventricular myocardium and right ventricle from MR images. Three datasets are used, ACDC\footnote{\url{https://www.creatis.insa-lyon.fr/Challenge/acdc/databases.html}}~\cite{Bernard_2018_ACDC}, {M$\&$Ms}\footnote{\url{https://www.ub.edu/mnms/}}~\cite{MMs_dataset_TMI_underreview} and corrupted ACDC, named as ACDC-C. For all experiments, the training set is a \textbf{single-site} set of only 10 subjects from ACDC. 10 and 20 subjects from ACDC are used for validation and intra-domain test. The multi-site M$\&$Ms dataset (150 subjects from 5 different sites) is used for cross-domain test. The ACDC-C dataset is used for evaluating the robustness of the method for corrupted images. Challenging scenarios are simulated, where 20 ACDC test subjects are augmented three times with four different types of MR artefacts: bias field, ghosting, motion and spike  artifacts~\cite{perezgarcia_torchio_2020} using the TorchIO\footnote{\url{https://github.com/fepegar/torchio}} toolkit. This produces 4 subsets with 60 subjects, named as
\emph{RandBias},
\emph{RandGhosting}, \emph{RandMotion},
 \emph{RandSpike} in experiments.

\noindent\textbf{Implementation and evaluation.}
We employed the image pre-processing and default DA pipeline described in~\cite{chen_2020_MICCAI}, including common photo-metric and geometric image transformations. 
Our encoder and decoder pairs support general structures. Without loss of generality, we used a U-net like structure\cite{Ronneberger_2015_unet}. Supple. Fig.5 visualizes detailed structures of  encoder-decoder pairs as well as the latent space decoupler. For mask generation, we randomly select one type of proposed masking schemes at training, where $p$ is randomly selected from [$0\%$,$50\%$].
We use the Adam optimizer with a batch size of 20 to update network parameters, with a learning rate=$1e^{-4}$. Our code will be available on the Github\footnote{\url{https://github.com/cherise215/Cooperative_Training_and_Latent_Space_Data_Augmentation}}. For all methods, we trained the same network \emph{three} times using a set of randomly selected 10 ACDC subjects (600 epochs each run, on an Nvidia$^{\tiny{\text{\textregistered}}}$, using Pytorch). The average Dice score is reported for segmentation performance evaluation.\\ 

\setlength\intextsep{0pt}
\begin{figure}[t]
    \centering
    \includegraphics[width=\textwidth]{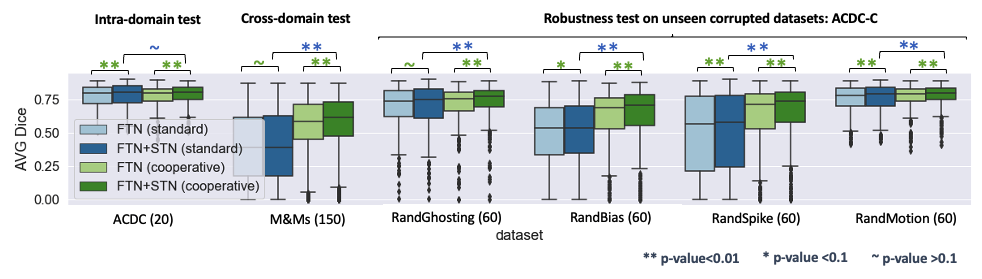}
    \caption{Compared to standard training, cooperative training with self-generating hard examples greatly improves the segmentation performance on various \emph{unseen, challenging} domains (p-value$<0.01$, average improvement: 15\%).}
    \label{fig:standard_vs_cooperative}
\end{figure}
\setlength\intextsep{0pt}
 \setlength\intextsep{0pt}
\begin{figure}[t]
    \centering
    \includegraphics[width=0.5\textwidth]{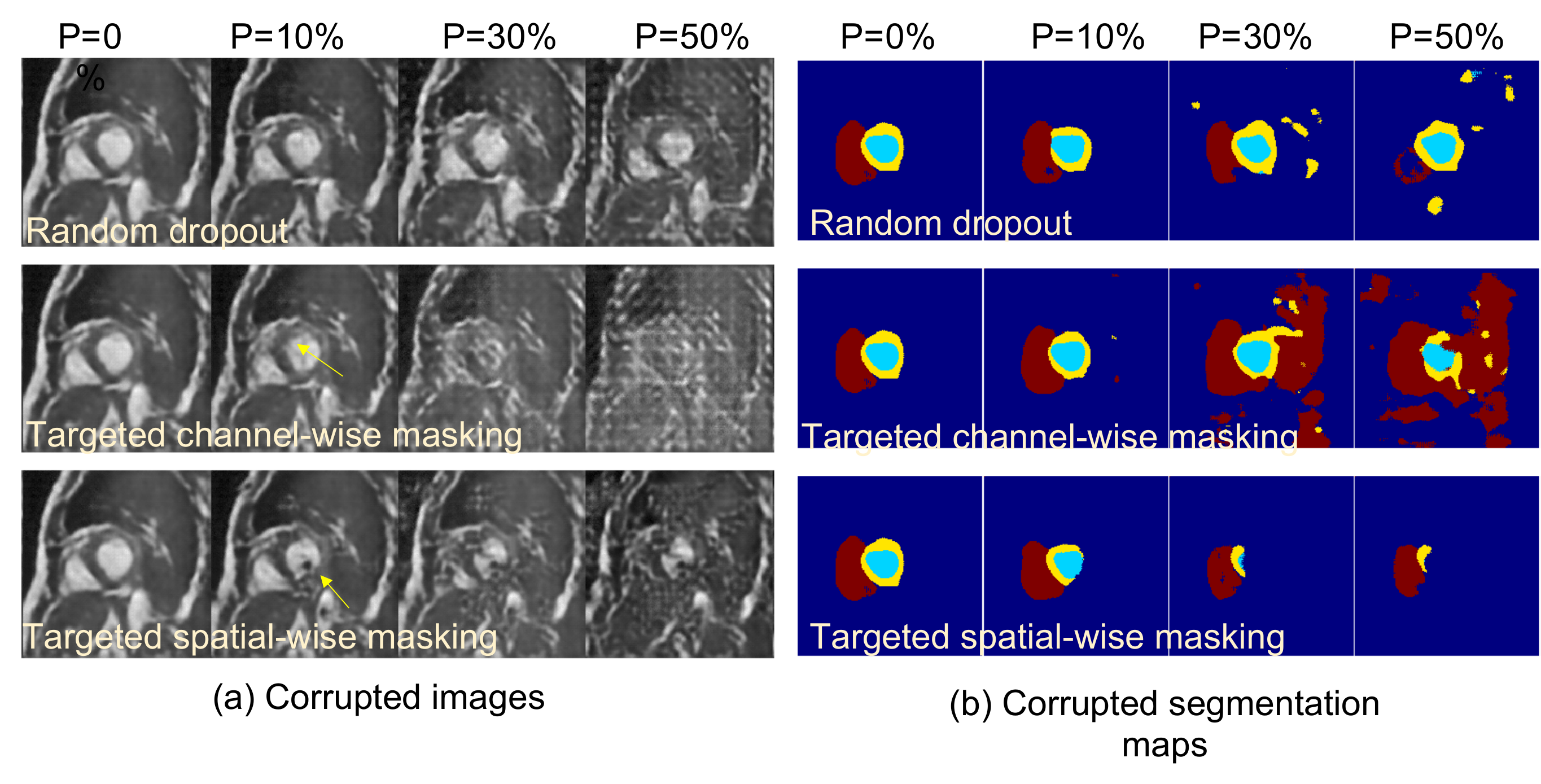}
    \caption{{Visualization of generated hard examples with the three masking schemes ($a$=0 for ease of comparison). Best viewed in zoom-in.}}
    \label{fig:hard_example}
\end{figure}
\setlength\intextsep{0pt}
 \setlength\intextsep{0pt}
\begin{table}[t]
\centering
\caption{Comparison to image space data augmentation methods for domain generalization. The proposed latent space augmentation method improves the performance on out-of-domain datasets compared to image space data augmentation methods. AVG: Average Dice scores across six datasets.}
\label{tab:SOTA_methods_vs_the_proposed}
\resizebox{\textwidth}{!}{\begin{tabular}{@{}lccccccc|c@{}}
\toprule
Method & ACDC & M\&Ms & RandBias & RandGhosting & RandMotion & RandSpike & \begin{tabular}[c]{@{}c@{}}AVG\\ (\FTN)\end{tabular} & \begin{tabular}[c]{@{}c@{}}AVG\\  (\FTN+\STN)\end{tabular} \\ \midrule
Standard training & 0.7681 & \textcolor{red}{0.3909} & \textcolor{red}{0.4889} & 0.6964 & 0.7494 & \textcolor{red}{0.4901} & 0.5970 & 0.6018 \\   \midrule 
Rand MWM~\cite{Zhou_2019_MICCAI_ModelsGenesis} & {0.7515} & \textcolor{red}{0.3984} & \textcolor{red}{0.4914} & 0.6685 & 0.7336 & {0.5713} & 0.6024 & 0.6131 \\
Rand Conv~\cite{Xu_2021_ICLR_RandConv} & {0.7604} & \textcolor{red}{0.4544} & 0.5538 & 0.6891 & 0.7493 & \textcolor{red}{0.4902} & 0.6162 & 0.6404 \\
Adv Noise~\cite{Miyato_2018_PAMI_VAT} & 0.7678 & \textcolor{red}{0.3873} & \textcolor{red}{0.4903} & 0.6829 & 0.7543 & {0.6244} & 0.6178 & 0.6276 \\
Adv Bias~\cite{chen_2020_MICCAI} & 0.7573 & \textbf{{0.6013}} & \textbf{{0.6709}} & 0.6773 & 0.7348 & \textcolor{red}{0.3840} & 0.6376 & 0.6604 \\ \midrule
Proposed w. $\hat\image$ & 0.7497 & 0.5154 & 0.5921 & 0.6921 & 0.7417 & \textbf{0.6633} & 0.6591 & 0.6709  \\
Proposed w. $\hat\image, \hat\prediction$ & \textbf{0.7696} & 0.5454 & 0.6174 & \textbf{0.7073} & \textbf{0.7643} & 0.6226 & \textbf{0.6711} & \textbf{0.6901} \\
\bottomrule
\end{tabular}}
\end{table}

\begin{table}[t]
\centering
\caption{Effectiveness of targeted masking, latent code decoupler $\filter$ and cooperative training}
\label{tab:ablation_joint_training_and_code_filteting}
\begin{adjustbox}{width=0.4\textwidth}
\begin{tabular}{@{}c|cc@{}}
\toprule
Methods & \FTN & \FTN+\STN \\ \midrule
w.o. $\gch,\gsp$ &  0.6344 & 0.6584 \\ \midrule
share code (a) ($\zi=\zi, \zs = \zi$) & 0.6625 & 0. 6868  \\
share code (b)  $(\zi=\zs, \zs=\zs)$ & {0.6343} & {0.6587} \\ \midrule
Separate Training~\cite{Larrazabal_2019_MICCAI_DAE}  & 0.6020 & 0.6077 \\ \midrule
Proposed & \textbf{0.6711} & \textbf{0.6901} \\ \bottomrule
\end{tabular}\end{adjustbox}
\end{table} 
\setlength\intextsep{0pt}
 \label{SEC: results}
\noindent\textbf{Experiment 1: Standard training vs cooperative training.} We compared the proposed cooperative training method with the standard training method (using $\Loss_{\mathit{standard}}$ only) using the same backbone network structure. Fig.~\ref{fig:standard_vs_cooperative} shows the box-plots for each method. While both methods achieve comparable performance on the intra-domain test set (p-value$>0.1$), it is clear that cooperative training with dual-network (\FTN+\STN) yields the best performance across out-of domain test sets (see dark green boxes). Consistent improvements made by \STN\ can be clearly observed across all domains. By contrast, \STN\ with standard training fails to provide significant improvements on some datasets (p-value$>0.1$). This indicates the superiority of cooperative training with latent space DA.\\

\noindent\textbf{Experiment 2: Latent space DA vs Image space DA.}
We compared the proposed latent space -based method to other competitive image space DA methods: a) \emph{random multi-window in-and-out masking (Rand MWM)}~\cite{Zhou_2019_MICCAI_ModelsGenesis,Zhou_2021_Media_ModelsGenesis}, which uses an enhanced variant of Cutout~\cite{Devries_2017_Arxiv_Cutout} and Patch Gaussian~\cite{Raphael_2019_Arxiv_PatchGaussian} to introduce patch-wise perturbation to images; b) \emph{random convolutional kernels (Rand Conv)}~\cite{Xu_2021_ICLR_RandConv}, which applies various random convolutional kernels to augment image texture and appearance variations; c) \emph{adversarial noise (Adv Noise)}~\cite{Miyato_2018_PAMI_VAT}; d) \emph{adversarial bias field (Adv Bias)}~\cite{chen_2020_MICCAI}, which augments image styles by adding realistic intensity inhomogeneities. We visualize augmented images using above methods in Supple. Fig.2 for better illustration.  For methods under comparison, we used their official code implementation if available and ran experiments using the same backbone network for fairness. Results are shown in Table~\ref{tab:SOTA_methods_vs_the_proposed}.

Surprisingly, with limited training data, both random and adversarial DA methods do not necessarily improve the network generalization on all datasets. While $Adv Bias$ achieves the best performance on \textit{M$\&$Ms} dataset and \textit{RandBias}, this method has a side effect, making it more sensitive to the spiking artifacts (Dice score 0.4901 vs 0.3840). By contrast, the proposed latent space DA achieves the top average performance across six datasets, without any dramatic failures (Dice score $<0.5$). Similar results can be found in a large training setting, see Supple. Fig.3. Our method can generate not only perturbed images but also realistically corrupted segmentations with increased uncertainty (Supple. Fig.4). These corrupted segmentations attribute to the increased model generalization (AVG Dice: 0.6709 vs 0.6901). 
While one may argue that characterizing and combining various image-space DAs and corruptions together could be an interesting direction to improve cross-domain performance, it is time-consuming and computationally inefficient to find the optimal DA policy~\cite{Cubuk_2019_CVPR_AutoAugment}, and has the risk of sacrificing intra-domain performance~\cite{Richard_2019_MIDL}.\\

\noindent\textbf{Experiment 3: Ablation study.} We further investigate three key contributions: 1) the proposed targeted masking; 2)
latent code decoupler $\filter$; 3) cooperative training. Results are shown in Table~\ref{tab:ablation_joint_training_and_code_filteting}.  We can see that disabling $\gch,\gsp$ drops the average Dice score from 0.6901 to 0.6584, highlighting the effectiveness of targeted masking. Fig.~\ref{fig:hard_example} shows that targeted masking focuses more on attacking cardiac structures, resulting in more challenging images with mixed artifacts and under or over-segmented predictions. We compared the proposed network architecture to its two variants, where $\zi$ and $\zs$ are shared in two different ways. Both variants lead to inferior performance. This suggests the benefit of $\filter$ for a more sparse $\zs$ code. Image reconstruction requires low-level information, whereas image segmentation relies on more concentrated high-level information. Introducing $\filter$ explicitly defines a hierarchical feature structure to improve model generalization. Lastly, we compared our method to the state-of-the-art denoising auto-encoder-based shape refinement method (Separate Training)~\cite{Larrazabal_2019_MICCAI_DAE} where \FTN\ and \STN\ are trained independently. It has been shown that this learning-based method can outperform the commonly used non-learning-based condition random field-based refinement method~\cite{Christ_2016_MICCAI}. Results show that our method can greatly outperform this advanced method by a large margin (Dice score 0.6901 vs. 0.6077), highlighting the benefits of the cooperative training strategy for enhancing learning-based shape refinement and correction.

 \section{Conclusion}
\label{SEC: conclusion}
 We present a novel cooperative training framework in together with a latent space masking-based DA method. Experiments show that it greatly improves model generalizability and robustness against unforeseen domain shifts. Unlike existing methods which require multi-domain datasets or domain knowledge to specify particular forms of image transformation and corruption functions, our latent space DA method requires \emph{little} human effort, and it has the potential to be applied to other data-driven applications. Although we only demonstrate the performance for cardiac image segmentation, our \emph{generic} framework has the potential to be extended to a wide range of data-driven applications.\\

\subsubsection{Acknowledgment:} This work was supported by the SmartHeart EPSRC Programme Grant(EP/P001009/1).
\bibliographystyle{unsrt}
\bibliography{mybib}

\begin{thebibliography}{10}

\bibitem{Shen_2017_Review}
Dinggang Shen et~al.
\newblock Deep learning in medical image analysis.
\newblock {\em Annual review of biomedical engineering}, 19:221--248, June
  2017.

\bibitem{litjens_2017_survey}
Geert Litjens et~al.
\newblock A survey on deep learning in medical image analysis.
\newblock {\em Medical image analysis}, 42:60--88, 2017.

\bibitem{Dou_2019_NIPS}
Qi~Dou et~al.
\newblock Domain generalization via model-agnostic learning of semantic
  features.
\newblock In Hanna~M. Wallach et~al., editors, {\em NeurIPS 2019}, pages
  6447--6458, 2019.

\bibitem{albuquerque_2020_Arxiv_improving}
Isabela Albuquerque et~al.
\newblock Improving out-of-distribution generalization via multi-task
  self-supervised pretraining.
\newblock {\em arXiv preprint arXiv:2003.13525}, 2020.

\bibitem{chattopadhyay_2020_ECCV_learning_to_balance}
Prithvijit Chattopadhyay et~al.
\newblock Learning to balance specificity and invariance for in and out of
  domain generalization.
\newblock In {\em European Conference on Computer Vision}, pages 301--318.
  Springer, 2020.

\bibitem{Wang_2020_ECCV}
Shujun Wang et~al.
\newblock Learning from extrinsic and intrinsic supervisions for domain
  generalization.
\newblock In {\em ECCV}, 2020.

\bibitem{Shankar_2018_ICLR}
Shiv Shankar et~al.
\newblock Generalizing across domains via cross-gradient training.
\newblock In {\em ICLR}. OpenReview.net, 2018.

\bibitem{daniel_2017_thinking}
Kahneman Daniel.
\newblock Thinking, fast and slow, 2017.

\bibitem{Shorten_2019_Bigdata}
Connor Shorten et~al.
\newblock A survey on image data augmentation for deep learning.
\newblock {\em Journal of Big Data}, 6(1):60, July 2019.

\bibitem{Zhang_2020_TMI}
Ling Zhang et~al.
\newblock Generalizing deep learning for medical image segmentation to unseen
  domains via deep stacked transformation.
\newblock {\em {IEEE} Trans. Medical Imaging}, 39(7):2531--2540, 2020.

\bibitem{chen_2020_frontiers}
Chen Chen et~al.
\newblock Improving the generalizability of convolutional neural network-based
  segmentation on cmr images.
\newblock {\em Frontiers in cardiovascular medicine}, 7:105, 2020.

\bibitem{Devries_2017_Cutout}
Terrance Devries et~al.
\newblock Improved regularization of convolutional neural networks with cutout.
\newblock {\em CoRR}, abs/1708.04552, 2017.

\bibitem{Raphael_2019_Arxiv_PatchGaussian}
Raphael~Gontijo Lopes et~al.
\newblock Improving robustness without sacrificing accuracy with patch gaussian
  augmentation.
\newblock {\em CoRR}, abs/1906.02611, 2019.

\bibitem{Zhou_2019_MICCAI_ModelsGenesis}
Zongwei Zhou et~al.
\newblock Models genesis: Generic autodidactic models for 3d medical image
  analysis.
\newblock In {\em MICCAI 2019}, pages 384--393, Cham, 2019. Springer
  International Publishing.

\bibitem{Zhou_2021_Media_ModelsGenesis}
Zongwei Zhou et~al.
\newblock Models genesis.
\newblock {\em Medical Image Analysis}, 67:101840, 2021.

\bibitem{Miyato_2018_PAMI_VAT}
Takeru Miyato, Shin-Ichi Maeda, Masanori Koyama, and Shin Ishii.
\newblock Virtual adversarial training: A regularization method for supervised
  and {Semi-Supervised} learning.
\newblock {\em TPAMI}, 2018.

\bibitem{chen_2020_MICCAI}
Chen Chen et~al.
\newblock Realistic adversarial data augmentation for mr image segmentation.
\newblock In {\em MICCAI}, pages 667--677. Springer, 2020.

\bibitem{Zhang_2020_DADA}
X.~{Zhang} et~al.
\newblock Deep adversarial data augmentation for extremely low data regimes.
\newblock {\em IEEE Transactions on Circuits and Systems for Video Technology},
  31(1):15--28, 2021.

\bibitem{Zhao_2020_NIPS}
Long Zhao et~al.
\newblock Maximum-entropy adversarial data augmentation for improved
  generalization and robustness.
\newblock In {\em NeurIPS}, 2020.

\bibitem{Zheng_2019_CVPR}
Wenzhao Zheng, Zhaodong Chen, Jiwen Lu, and Jie Zhou.
\newblock Hardness-aware deep metric learning.
\newblock In {\em CVPR}, pages 72--81, 2019.

\bibitem{zhang_2017_survey_MTL}
Yu~Zhang et~al.
\newblock A survey on multi-task learning.
\newblock {\em arXiv preprint arXiv:1707.08114}, 2017.

\bibitem{Larrazabal_2019_MICCAI_DAE}
Agostina~J Larrazabal et~al.
\newblock Anatomical priors for image segmentation via post-processing with
  denoising autoencoders.
\newblock In {\em {MICCAI} 2019}, pages 585--593. Springer International
  Publishing, 2019.

\bibitem{painchaud_2019_MICCAI_cardiac}
Nathan Painchaud et~al.
\newblock Cardiac mri segmentation with strong anatomical guarantees.
\newblock In {\em MICCAI}, pages 632--640. Springer, 2019.

\bibitem{Tishby_2000_information}
Naftali Tishby et~al.
\newblock The information bottleneck method.
\newblock {\em arXiv preprint physics/0004057}, 2000.

\bibitem{Huang_2020_ECCV_Self_challenging}
Zeyi Huang et~al.
\newblock Self-challenging improves cross-domain generalization.
\newblock In Andrea Vedaldi, Horst Bischof, Thomas Brox, and Jan{-}Michael
  Frahm, editors, {\em ECCV}, volume 12347 of {\em Lecture Notes in Computer
  Science}, pages 124--140. Springer, 2020.

\bibitem{Tompson_2015_CVPR_SpatialDropout}
Jonathan Tompson et~al.
\newblock Efficient object localization using convolutional networks.
\newblock In {\em CVPR}, pages 648--656. {IEEE} Computer Society, 2015.

\bibitem{Madry_2017_PGDattack}
Aleksander Madry, Aleksandar Makelov, Ludwig Schmidt, Dimitris Tsipras, and
  Adrian Vladu.
\newblock Towards deep learning models resistant to adversarial attacks.
\newblock In {\em {ICLR}}, June 2017.

\bibitem{Goodfellow_2015_FGSM}
Ian~J Goodfellow et~al.
\newblock Explaining and harnessing adversarial examples.
\newblock In {\em {ICLR}}, 2015.

\bibitem{Bernard_2018_ACDC}
Olivier Bernard et~al.
\newblock Deep learning techniques for automatic {MRI} cardiac
  {Multi-Structures} segmentation and diagnosis: Is the problem solved?
\newblock {\em TMI}, 0062(11):2514--2525, November 2018.

\bibitem{MMs_dataset_TMI_underreview}
Víctor~M. Campello et~al.
\newblock Multi-centre, multi-vendor and multi-disease cardiac segmentation:
  The m\&ms challenge. ieee transactions on medical imaging, under review.
\newblock {\em TMI}.

\bibitem{perezgarcia_torchio_2020}
Fernando P{\'e}rez-Garc{\'i}a et~al.
\newblock {TorchIO}: a {Python} library for efficient loading, preprocessing,
  augmentation and patch-based sampling of medical images in deep learning.
\newblock {\em arXiv:2003.04696 [cs, eess, stat]}, March 2020.

\bibitem{Ronneberger_2015_unet}
Olaf Ronneberger et~al.
\newblock {U-Net}: Convolutional networks for biomedical image segmentation.
\newblock In {\em {MICCAI} 2015}, volume 9351 of {\em Lecture Notes in Computer
  Science}, pages 234--241. Cham, May 2015.

\bibitem{Xu_2021_ICLR_RandConv}
Zhenlin Xu et~al.
\newblock Robust and generalizable visual representation learning via random
  convolutions.
\newblock In {\em ICLR}, 2021.

\bibitem{Devries_2017_Arxiv_Cutout}
Terrance Devries et~al.
\newblock Improved regularization of convolutional neural networks with cutout.
\newblock {\em CoRR}, abs/1708.04552, 2017.

\bibitem{Cubuk_2019_CVPR_AutoAugment}
Ekin~D Cubuk et~al.
\newblock Autoaugment: Learning augmentation strategies from data.
\newblock In {\em CVPR}, pages 113--123, 2019.

\bibitem{Richard_2019_MIDL}
Richard Shaw et~al.
\newblock Mri k-space motion artefact augmentation: Model robustness and
  task-specific uncertainty.
\newblock In M.~Jorge Cardoso et~al., editors, {\em MIDL}, volume 102 of {\em
  Proceedings of Machine Learning Research}, pages 427--436, London, United
  Kingdom, 08--10 Jul 2019. PMLR.

\bibitem{Christ_2016_MICCAI}
Patrick~Ferdinand Christ et~al.
\newblock Automatic liver and lesion segmentation in {CT} using cascaded fully
  convolutional neural networks and {3D} conditional random fields.
\newblock In {\em MICCAI 2016}, pages 415--423. Springer International
  Publishing, 2016.

\end{thebibliography}
\clearpage
\appendix
\section{Supplementary Material}


\setcounter{page}{1}
\setcounter{figure}{0}
\renewcommand{\figurename}{Supple. Figure}
\setcounter{table}{0}
\renewcommand{\tablename}{Supple. Table}



\begin{figure}[!ht]
\centering
 \begin{subfigure}[b]{0.49\textwidth}
 \centering
 \includegraphics[width=\textwidth]{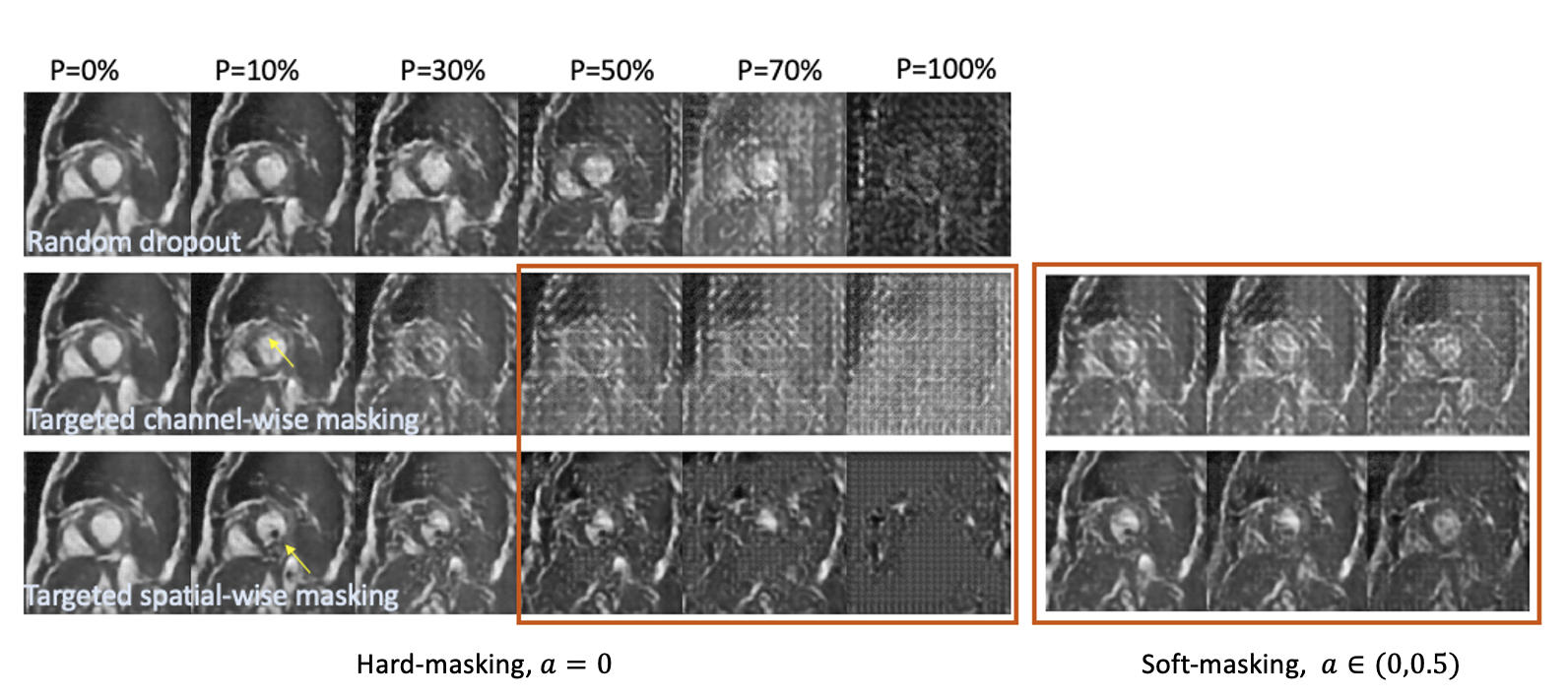}
 \caption{Corrupted images}
 \label{fig:Generated corrupted image}
 \end{subfigure}
 \hfill
\begin{subfigure}[b]{0.49\textwidth}
         \centering
         \includegraphics[width=\textwidth]{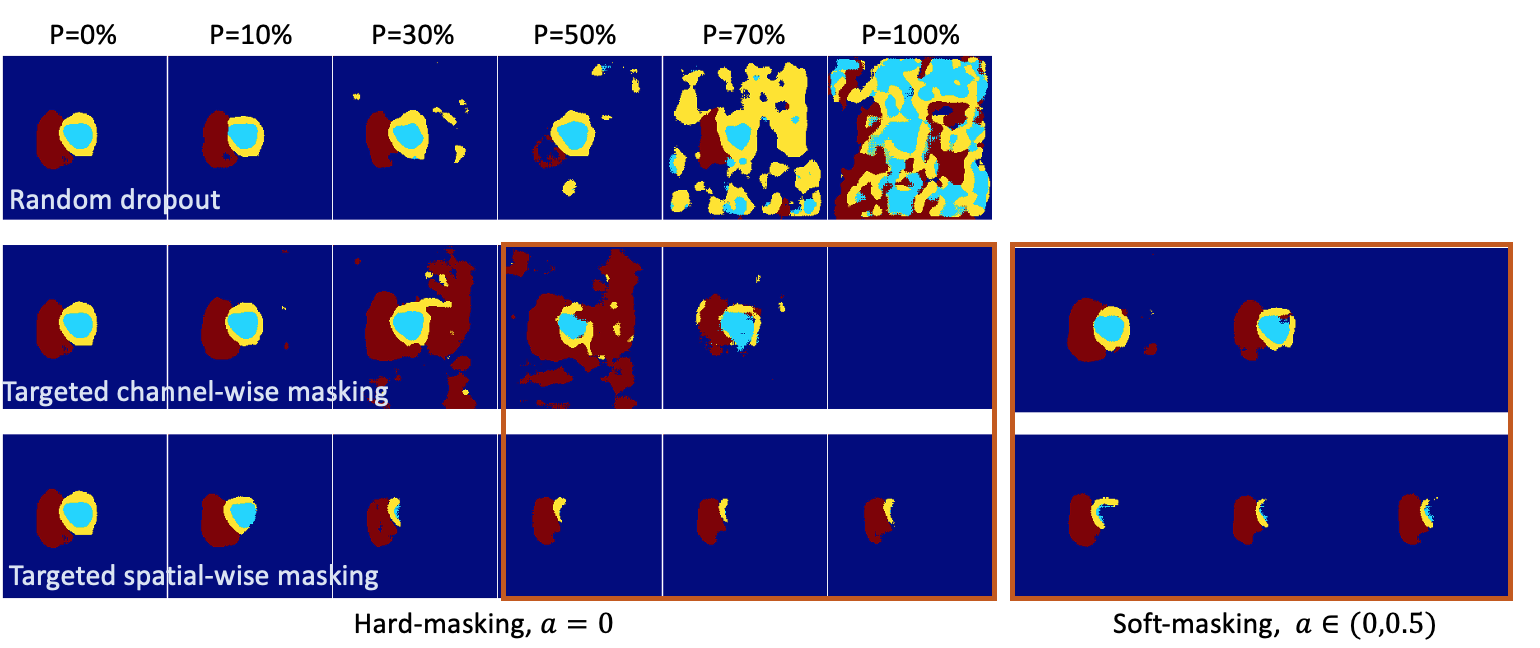}
         \caption{Corrupted segmentations}
         \label{fig:Generated corrupted segmentation}
\end{subfigure}
\caption{Three types of latent code masking schemes generate (a) a \emph{diverse} set of challenging images with \emph{unseen} mixed artifacts, e.g. `dark dots', `checkerboard artifacts', `blurring' and (b) various over-segmented and under-segmented predictions at different thresholds $p$. Compared to hard-masking, soft-masking produces milder but more \emph{diverse} corrupted images and segmentation maps. a: the annealing factor for soft masking.}
\end{figure}

\setlength\intextsep{0pt}
\begin{figure}[!ht]
    \centering
    \includegraphics[width=0.9\textwidth]{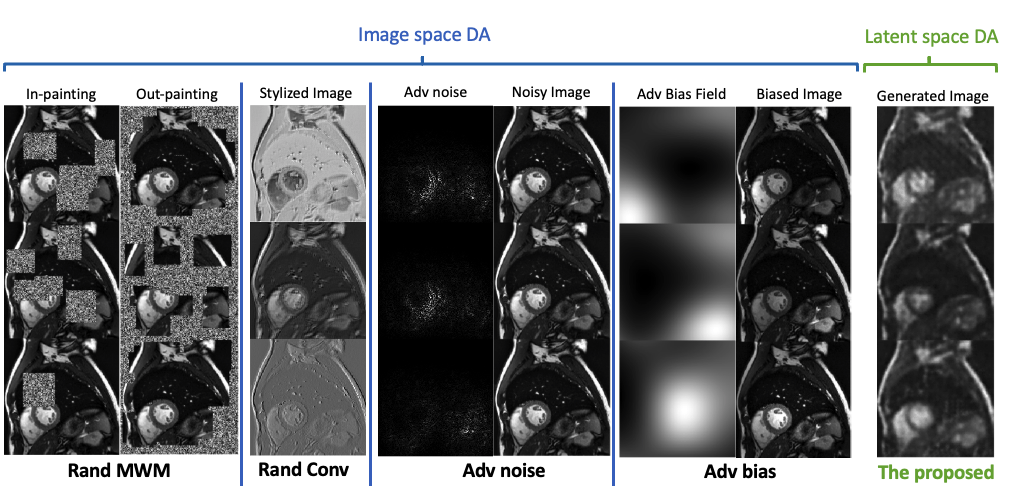}
    \caption{Visualization of input space data augmentation and latent space data augmentation (ours). DA: data augmentation. Adv: Adversarial.}
    \label{fig:image_vs_latent}
\end{figure}
\setlength\intextsep{0pt}

\setlength\intextsep{0pt}
\begin{figure}[!ht]
    \centering
    \includegraphics[width=0.9\textwidth]{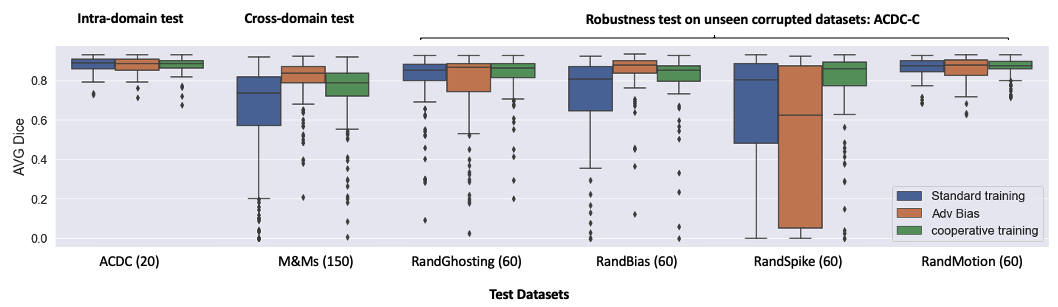}
    \caption{In the large training data setting (70 ACDC subjects), when compared to the baseline method (standard training), our cooperative training method can further improve not only intra-domain segmentation accuracy (with reduced variance) but also  robustness against various domain shifts. Adv bias, by contrast, fails to provide consistent improvement. This reveals our method's great potential to be applied to a wide range of scenarios for both improved generalization and robustness.}
    \label{fig:adv_bais_vs_cooperative_large}
\end{figure}
\setlength\intextsep{0pt}


\setlength\intextsep{0pt}
\begin{figure}[!ht]
\begin{minipage}{.48\textwidth}
 \centering
    \includegraphics[width=\textwidth]{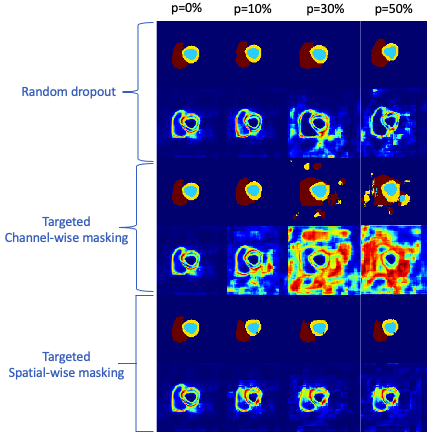}
    \caption{Visualization of corrupted segmentation (the first row in each block) and the corresponding entropy map (the second row in each block) using the proposed three latent space masking schemes. Latent masking schemes generate \textbf{realistic} poor segmentation with \textbf{increased} entropy, which is beneficial to train our denoising autoencoder (STN) for shape correction.}
    \label{fig:segmentation_and_entropy}
\end{minipage}
 \hfill
\begin{minipage}{.5\textwidth}
  \centering
    \includegraphics[width=\textwidth]{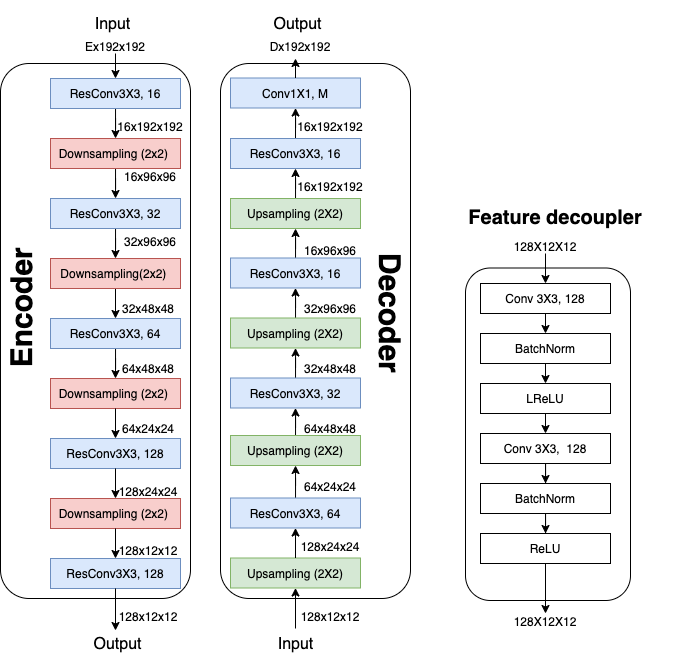}
    \caption{{Structures of Unet-like encoder-decoder pairs, and the feature decoupler used in this paper. We use the same structures for encoders and decoders accordingly. E: \# of input channel(s), D: \# of output channel(s). ResConv: Convolutional Block with residual connections~\cite{he_2016_identity}. Conv: Standard convolutional kernels. Of note, our framework is \textbf{generic}, other encoders and decoders can also be used.}}
    \label{fig:network}
\end{minipage}
\end{figure}
\setlength\intextsep{0pt}

\end{document}